\def\year{2021}\relax
\documentclass[letterpaper]{article} 
\usepackage{aaai21}  
\usepackage{times}  
\usepackage{helvet} 
\usepackage{courier}  
\usepackage[hyphens]{url}  
\usepackage{graphicx} 
\usepackage{natbib}  
\usepackage{caption} 
\usepackage{subcaption,booktabs}
\usepackage{amsmath, mathtools, amsfonts}
\usepackage{color}

\title{Adaptive Agent Architecture for Real-time Human-Agent Teaming}
\author{
    Tianwei Ni\textsuperscript{\rm 1}\thanks{indicates equal contribution. In AAAI 2021 Workshop on Plan, Activity, and Intent Recognition.},
    Huao Li\textsuperscript{\rm 2}\footnotemark[1],
    Siddharth Agrawal\textsuperscript{\rm 1}\footnotemark[1],
    Suhas Raja\textsuperscript{\rm 3},
    Fan Jia\textsuperscript{\rm 1},
    Yikang Gui\textsuperscript{\rm 2}, \\
    Dana Hughes\textsuperscript{\rm 1},
    Michael Lewis\textsuperscript{\rm 2},
    Katia Sycara\textsuperscript{\rm 1}\\
}
\affiliations{
    \textsuperscript{\rm 1}Carnegie Mellon University, 
    \textsuperscript{\rm 2}University of Pittsburgh,
    \textsuperscript{\rm 3}University of Texas at Austin\\ 
    tianwein@cs.cmu.edu, hul52@pitt.edu, siddhara@cs.cmu.edu
}

\begin{document}

\maketitle


\newcommand{\tw}[1]{\textcolor{blue}{TN: #1}}
\newcommand{\sr}[1]{\textcolor{red}{SR: #1}}
\newcommand{\sa}[1]{\textcolor{green}{SA: #1}}
\newcommand{\hl}[1]{\textcolor{purple}{HL: #1}}

\newcommand{\E}[2]{\mathbb{E}_{#1}{\left[#2\right]}}
\newcommand*{\argmax}{\mathop{\mathrm{argmax}}}
\newcommand*{\argmin}{\mathop{\mathrm{argmin}}}
\newcommand{\defeq}{\mathrel{\mathop:}=}

\newcommand{\B}{\mathcal B}
\renewcommand{\S}{\mathcal S}

\begin{abstract}
Teamwork is a set of interrelated reasoning, actions and behaviors of team members that facilitate common objectives. Teamwork theory and experiments have resulted in a set of states and processes for team effectiveness in both human-human and agent-agent teams.
However, human-agent teaming is less well studied because it is so new and involves asymmetry in policy and intent not present in human teams. To optimize team performance in human-agent teaming, it is critical that agents infer human intent and adapt their polices for smooth coordination. 
Most literature in human-agent teaming builds agents referencing a learned human model. Though these agents are guaranteed to perform well with the learned model, they lay heavy assumptions on human policy such as optimality and consistency, which is unlikely in many real-world scenarios.
In this paper, we propose a novel adaptive agent architecture in human-model-free setting on a two-player cooperative game, namely Team Space Fortress (TSF). Previous human-human team research have shown complementary policies in TSF game and diversity in human players' skill, which encourages us to relax the assumptions on human policy.
Therefore, we discard learning human models from human data, and instead use an adaptation strategy on a pre-trained library of exemplar policies  composed of RL algorithms or rule-based methods with minimal assumptions of human behavior.
The adaptation strategy relies on a novel similarity metric to infer human policy and then selects the most complementary policy in our library to maximize the team performance. 
The adaptive agent architecture can be deployed in real-time and generalize to any off-the-shelf static agents.
We conducted human-agent experiments to evaluate the proposed adaptive agent framework, and demonstrated the suboptimality, diversity, and adaptability of human policies in human-agent teams.
\end{abstract}

\section{Introduction}

Multi-agent systems have recently seen tremendous progress in teams of purely artificial agents, especially in computer games~\cite{vinyals2019grandmaster,guss2019minerl,openai2019dota}. 
However, many real-world scenarios like autonomous driving~\cite{sadigh2018planning, fisac2019hierarchical}, assisted robots~\cite{agrawal2017robot,li2019perceptions}, and Unmanned Aerial System~\cite{mcneese2018teaming,demir2017team} do not guarantee teams of homogeneous robots with shared information - more often, it involves interaction with different kinds of humans who may have varying and unknown intents and beliefs. Understanding these intents and beliefs is crucial for robots to interact with humans effectively in this scenario.
\textbf{Human-agent teaming} (HAT)~\cite{scholtz2003theory,chen2014human}, an emerging form of human-agent systems, requires teamwork to be a set of interrelated reasoning, actions and behaviors of team members that combine to fulfill team objectives~\cite{morgan1986measurement,salas2005there,salas2008teams}.
In this paper, we focus on the setting of two-player human-agent teaming in a computer game, where the agent should cooperate with the human in real-time to achieve a common goal on one task. The human playing that role may be any person with any policy at any time, and potentially not be an expert in the task at hand.

One of the fundamental challenges for an artificial agent to work with a human, instead of simply another artificial agent, is that humans may have complex or unpredictable behavioral patterns and intent~\cite{chen2012supervisory,green2006enumeration}. In particular, they may misuse or disuse the multi-agent system based on their perception, attitude and trust towards the system~\cite{parasuraman1997humans}. 
This difference becomes very critical in scenarios where an agent interacts with a diverse population of human players, each of which might have different intents, beliefs, and skills (ranging from novices to experts)~\cite{kurin2017atari}.
To succeed, cooperative agents must be able to infer human intent or policy to inform their action accordingly.

Capabilities of adapting to humans are essential for a human-agent team to safely deploy and collaborate in a real-time environment~\cite{bradshaw2011human}. Real-time adaptation is critical in practical deployments where robots are not operating unilaterally in a controlled environment, such as urban driving environments for autonomous vehicles~\cite{fisac2019hierarchical}. The ability to respond to real-time observations improves an agent’s ability to perform in the face of various kinds of team structures or situations. 
Accomplishing real-time agent adaptation requires that agents are able to capture the semantics of the observed human behavior, which is likely volatile and noisy, and then infer a best response accordingly. The challenge of capturing human behavior is further increased since the agent only observes a small snapshot of recent human actions, among players with varying play styles or skill levels. 
Finally, we note that humans may adjust their behavior in response to a changing agent policy, which can make stable adaptation difficult to achieve~\cite{haynes1996co,fisac2019hierarchical,sadigh2016information}. Real-time environments like computer games also require agents to perform both sufficiently fast estimation of the teammate's policy, as well as planning, while ensuring flexibility for unexpected strategic game states~\cite{vinyals2017starcraft}.

Past research on real-time adaptation in HAT can be divided into two forms of adaptive agent training. 
In the first \textit{human-model-free} form: the agent does not build a model of human policy, but instead infers human types from the match between current observations and an exemplar and then take corresponding best actions.
This setting is adopted by our approach and can be found in the psychology literature~\cite{kozlowski2018unpacking,kozlowski2015teams,kozlowski2000multilevel}.
The second form widely adopted in robotics research  \textit{human-model-based}: first trains a model of the human to learn humans' policies or intent, then integrates the human model into the environment, and finally trains the agent upon the integrated environment. 
This setting requires much more computational resources than \textit{human-model-free} form to learn the human model and deploy the agent in real-time, and inevitably imposes heavy assumptions on human policies~\cite{zakershahrak2018interactive, sadigh2018planning,fisac2019hierarchical} like: optimal in some unknown reward function, single-type or consistent among different humans, and time-invariant for one human, etc. 
However these assumptions deviate from real-world human policies, especially coming from a diverse population in different skill levels and intent on a relatively hard task. 
On the contrary, \textit{human-model-free} setting imposes minimal assumptions on human policies and can be deployed in real-time teaming efficiently.

In this paper, we propose a human-model-free adaptive agent architecture based on a pre-trained static agent library. The adaptive agent aims to perform well in a nontrivial real-time strategic game, Team Space Fortress (TSF)~\cite{agarwal2018challenges}. 
TSF is a two-player cooperative computer game where the players control spaceships to destroy the fortress. TSF presents a promising arena to test intelligent agents in teams since it involves heterogeneous team members (bait and shooter) with adversary (fortress), and it has sparse rewards which makes model training even more difficult.
TSF is a nontrivial testbed to solve as it requires some real-time cooperation strategy for the two players without communication and control skills for human players. 
Before constructing exemplar policies, we first evaluate the nature of this testbed through previous research in \textit{human-human} teams. The results~\cite{li2020team} show that different human-human pairs demonstrate significantly diverse performance and the team performance was affected by both individual level factors such as skill levels and team level factors such as team synchronization and adaptation.

The diverse team performance and complicated team dynamics in \textit{human-human} teams inspired us to build a real-time adaptive agent to cooperate with any human player in \textit{human-agent} teams. The methodology of our real-time adaptive agent is quite straightforward.
First, we design a diverse policy library of rule-based and reinforcement learning (RL) agents that can perform reasonably well in TSF when paired with each other, i.e. \textit{agent-agent} teams. We record the self-play performance of each pair in advance.
Second, we propose a novel similarity metric between any human policy and each policy in the library from observed human behavior, namely cross-entropy method (CEM) adapted from behavior cloning~\cite{bain1995framework}. 
The adaptive agent uses the similarity metric to find the most similar policy in the exemplar policies library to the the current human trajectory. After this, the adaptive agent switches its policy to the best complementary policy to the predicted human policy in real-time.
Using this adaptive strategy, it is expected to outperform any static policy from the library. Our approach is directly built upon single-agent models, thus can generalize to any off-the-shelf reinforcement learning/imitation learning algorithms.

We evaluated our approach online by having human players play against both agents with exemplar static policy and adaptive policy. These players were sourced through Amazon's Mechanical Turk (MTurk)\footnote{\url{https://www.mturk.com/}} program and played TSF through their internet browsers.
Each human player was assigned one role in TSF and played with all the selected agents for several trials, but was not told which agents they were playing with and was rotated through random sequences of the agents to ensure agent anonymity and reduce learning effect.

Based on the collected game data from these human-agent teams, we are interested in the three key questions: 
(1)How are human players' policies compared to agent policies in our library? (2) Is our adaptive agent architecture capable of identifying human policy and predicting team performance for human-agent teams? (3) Do our adaptive agents perform better than static policy agents in human-agent teams? We answer these questions in the experimental section.

\section{Related Work}
In the multi-agent system domain, researchers have been focusing on how autonomous agents model other agents in order to better cooperate with each other in teams, which is termed as \textit{human-model-based} methods when applied to human-agent system in the introduction of this paper. Representative work includes \textit{ad-hoc teamwork} that the agent is able to use prior knowledge about other teammates to cooperate with new unknown teammates~\cite{barrett2015cooperating,albrecht2018autonomous}.

This is a reasonable number of work in human-robot interaction that attempts to infer human intent from observed behaviors using inverse planing or inverse reinforcement learning
~\cite{bajcsy2018learning, sadigh2017active, sadigh2018planning, reddy2018you, fisac2019hierarchical}. However, these work impose ideal assumptions on human policy, e.g. optimal under some unknown reward, consistent through time, and with unique type among humans, which does not hold in many complicated real-world applications, where the human-agent systems are required to generalize to various kinds of team scenarios.

In human-agent teaming, past research~\cite{fan2009human,harbers2012enhancing, van2020learning,levine2018watching, chen2018situation} has established a variety of protocols within small teams. However, these approaches often rely on some degree of explicit communication on humans' observation or intent.

The alternative setting of agent design in human-agent system is \textit{human-model-free}, rarely discussed in the robotics literature. 
Some psychology literature in this setting learns to infer human intent from retrospective teammate reports, where software analyzes historical observations of humans to inform behavior in the present~\cite{kozlowski2018unpacking,kozlowski2015teams,kozlowski2000multilevel}. These historical behaviors may fail to capture potential changes in teammate policies a real-time environment and limit the ability of software to best adapt to a situation.

Our approach opens the door of \textit{human-model-free} setting in human-agent system for robotics literature, significantly different from previous \textit{human-model-based} methods. 
We make least assumptions on human, and use proposed architecture to realize adaptation, which involves similarity metric to infer human policy. 
The least assumptions and straightforward architecture enable our approach to deploy in a real-time human-agent environment with various kinds of human players.

\section{Team Space Fortress}

We have adapted Space Fortress~\cite{mane1989space}, a game which has been used extensively for psychological research, for teams. Team Space Fortress (TSF) is a cooperative computer game where two players control their spaceships to destroy a fortress in a low friction 2D environment. The player can be either human or (artificial) agent, thus there are three possible combinations in teams: human-human, human-agent, and agent-agent.

\begin{figure}[h]
    \centering
    \includegraphics[width=0.32\textwidth]{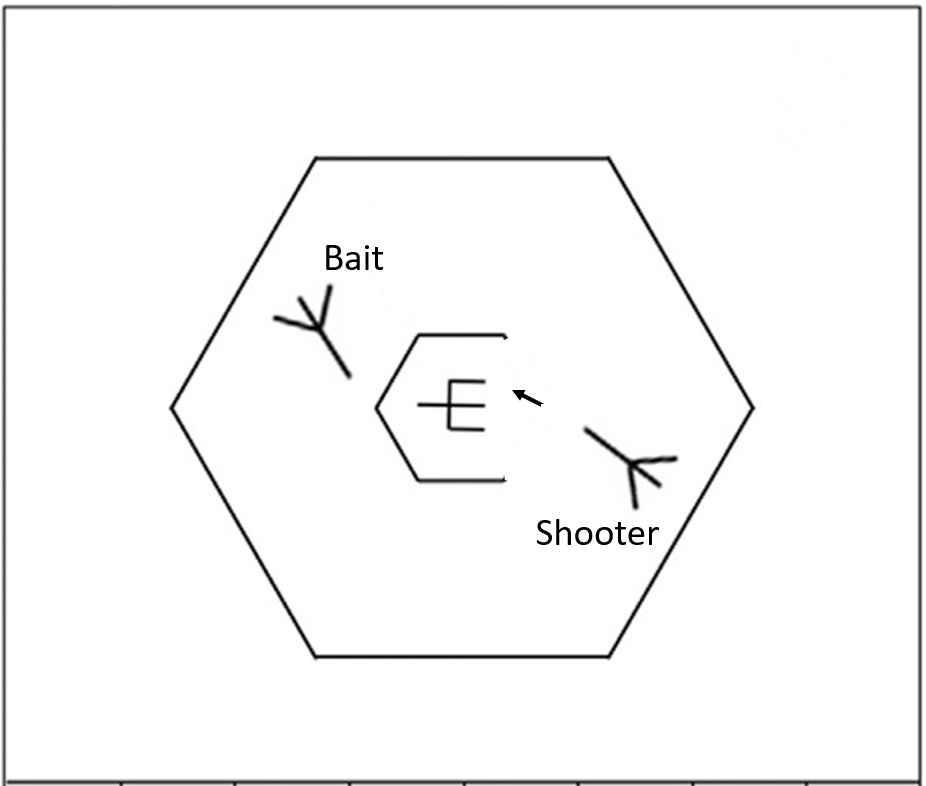}
    \caption{\footnotesize Sample TSF game screen (line drawing version, original screen is in black background). Spaceships are labeled as shooter and bait. Entity at center is the rotating fortress with the boarder around it as the shield. Activation region is the hexagon area around players' spaceships. Black arrow is a projectile emitted from the shooter towards the fortress. All the entities are within the rectangle map borders.}
    \label{fig:TSF}
\end{figure}

A sample screen from the game is shown in Fig. \ref{fig:TSF}. 
At the center of the stage lies a rotating fortress. The fortress and two spaceships can all fire missiles towards each other at a range. The first spaceship entering the hexagon area will be locked and shot by the fortress. However, the fortress becomes vulnerable when it is firing. Players die immediately whenever they hit any obstacles (e.g. boundaries, missiles, the fortress). The game resets every time either fortress or both players are killed. Once the fortress has been killed, both players must leave the activation region (outer pink boundaries) before the fortress respawns. 

The team performance is measured by the number of fortresses that players kill. 
The action space is 3-dimensional discrete space, including TURN (turn left, right, or no turn), THRUST (accelerate the speed or not), and FIRE (emit one missile or not).
The frame per second (FPS) is 30, thus in a 1-minute game, there are around 1800 frames.

In order to test a common instance of teamwork, players were instructed in a common strategy and assigned roles of either \textit{bait} or \textit{shooter}.
The \textit{bait} tries to attract the fortress's attention by entering the inner hexagon where it is vulnerable to the fortress. When the fortress attempts to shoot at the bait, its shield lifts making it vulnerable. The other player in the role of \textit{shooter} can now shoot at the fortress and destroy it. 

There are some difference in observations and actions between human players and agent players.
Human players observe the game screen (RGB image) at each frame, then hit or release the keys on keyboard to take actions. 
Agent players instead observe an array composed of the states (position, velocity, angle) of all the entities including the fortress and two players, and communicate their actions directly through the game engine.

\section{Adaptive Agent Architecture}
In this section, we formulate our method for an adaptive agent architecture. 
First we introduce the exemplar policies library pre-trained by reinforcement learning or designed by rules for TSF. This library will be used as a standard baseline to identify human policies. 
Next, we introduce the similarity metric adopted in the architecture (i.e. cross-entropy metric) that measures the distance between human trajectory and exemplar policies in the library.
Finally, we define the adaptive agent architecture given the estimated human policy according to the similarity metric.

\begin{figure}[h]
    \centering
    \includegraphics[width=0.45\textwidth]{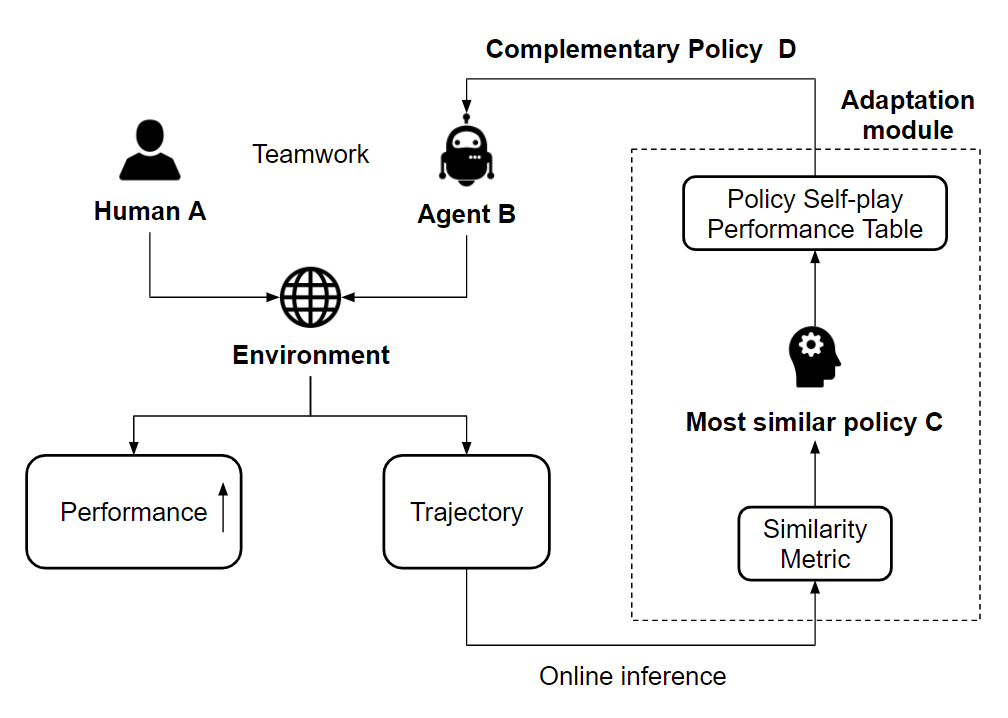}
    \caption{\footnotesize The flowchart of the proposed adaptive agent architecture. The adaptation module (in dotted boarder) takes the input of the trajectory at current timestamp, and then assigns the adaptive agent with new policy at next timestamp. The adaption procedure can be deployed in real-time (online).}
    \label{fig:flowchart}
\end{figure}

\subsection{Exemplar Policies Library}
\label{sec:lib}
The exemplar policies library $\mathcal L = \mathcal L_\B \cup \mathcal L_\S$ consists of two sets of policies in bait ($\B$) and shooter ($\S$) roles, $\mathcal L_\B$ and $\mathcal L_\S$ respectively. 
Both bait policies and shooter policies are trained using a combination of RL and rule based behavior. 

These exemplar policies can be divided into several main types: baits can be divided into three types (B1-B3, B4-B7, B8-B9) and shooters can be divided into two types (S1-S3, S4-S7).
Below are the technical details of each type.

\subsubsection{Bait policy library $\mathcal L_\B$}
To make these different bait policies diverse, we train them using different reward functions, inspired from human-human experiments where there were multiple ways to achieve good performance.
The reward functions attempt to encode the desirable behavior of a bait agent. The bait agent is then trained using an RL algorithm to achieve an optimal behavior with respect to the given reward function.
The bait library $\mathcal L_\B$ are composed of 9 bait policies.
The goal for bait is to keep alive inside the activation region to make the fortress vulnerable from behind so that the shooter can grasp the opportunity to destroy the fortress from behind. 
In general, the bait has two \textit{conflicting} objectives which it tries to balance. If the bait is inside the activation more time, it is more vulnerable and prone to getting killed by the fortress. 
However, bait's presence inside the activation region gives an opportunity for the shooter to attack the fortress from behind. 
Different bait agents try to balance these two conflicting objectives in different ways.

B1-B3 type policies are trained by A2C algorithm~\cite{konda2000actor} to learn TURN action and use rules on THRUST action. The reward function of TURN learning is binary, encouraging the baits to stay inside the activation region. 
For the observation space of the bait policy, we convert the original Cartesian coordinate system to a new one, where the new origin is still at the fortress while the new positive Y-axis goes through the bait, which is shown to ease the training in RL for TSF. Then, we use the converted coordinates of bait position and two nearest missile positions to train the agent. The intuition is that bait is sensitive to the nearest shells to keep itself alive. The rule in THRUST action limits the maximum speed of the bait agents. By tuning the threshold in speed, we have policy B1-B3.
B4-B7 policies are trained by RL in both TURN and THRUST actions. They share same reward function as B1-B3 using the same transformation in coordinate system. By using different RL algorithms from A2C~\cite{konda2000actor}, PPO~\cite{schulman2017proximal}, to TRPO~\cite{schulman2015trust}, and different observation space (whether to perceive the shooter's position and velocity), we have policies B4-B7.

B8-B9 belong to the another set of policy using a different reward structure, learning TURN and THRUST by PPO algorithm~\cite{schulman2017proximal}. B8 and B9 are designed by aggressive and defensive objectives, respectively.
The reward structure used to train these 2 agents is composed of three parts: (1) ``border reward" to encourage the agents to keep far away from the fortress, (2) ``bearing reward" to encourage the agents to align itself directly towards the fortress when inside the activation region, (3) ``death penalty" to discourage the agents from being killed by the fortress or hitting the border. Border reward encourages risk-averse behavior, while bearing reward risk-seeking behavior, and by controlling the coefficients among these three parts, we have agents B8-B9.

\subsubsection{Shooter policy library $\mathcal L_\S$}

The shooter policy library $\mathcal L_\S$ are composed of 7 shooter policies, with 4 of them mirror shooters that are purely rule-based, and 3 of them RL shooters that learn the TURN action by RL.

The mirror shooters are based on the prior knowledge of TSF game that a good shooter should have an \textit{opposite} position against the Bait, which was observed in many successful human-human teams.
Thus the mirror shooter tries to keep at opposite position to the current position of the bait (termed as \textit{target position}) until it finds itself having a good chance to fire to destroy the fortress. By controlling the threshold of distance to the target position, we have agents S4-7.

The RL shooters' reward function takes the same team strategy of opposite positions, trained by DDQN~\cite{van2015deep} on TURN action.
Specifically, the reward is designed to encourage the shooter to keep close to the outside the activation region when bait does not enter the region, and keep at the rear of the fortress when bait enters the region. S1-3 are different in max speed.

\subsubsection{Self-play performance}
\label{sec:self-play}
We evaluate the performance of each shooter-bait pairs in the exemplar policy library by self-play in TSF environment, and record the results in self-play performance table $\mathcal P$ in advance. 
The table $\mathcal P$ has rows with the number of bait policies in $\mathcal L_\B$ and columns with the number of shooter policies in $\mathcal L_\S$, with each entry the average performance of the bait-shooter pair. When applied to our policy library, table $\mathcal P$ is showed in Table~\ref{tab:perf}.

On average, teams that consist of two static agents show significantly better performance ($6.03$) than human-human teams ($2.60$) reported in previous research~\cite{li2020team}. This indicates that most of our agent pairs, including both RL-based and rule-based agents, have a super-human performance in TSF which benefits from the design of reward function and rules. 


Similar to human-human teams, agent-agent teams also show complementary policy pairs that work extremely well with each other. An example would be S4-S7 (mirror shooters) who yield a dominant performance when pairing with most of the baits except for B8 and B9. While for specific bait policies such as B8 and B9, the best teammate would be S2 or S3 (RL shooters) in stead of the more ``optimal" S4-S7.
We could tell from the self-play table that the space of reasonable policies in TSF game is indeed diverse, and there are more than one path towards good team dynamics and team performance. This confirms again the necessity of introducing real-time adaptive agents in human-agent teams. Thus we build the adaptive agent framework based upon this self-play table in the next subsections.

\begin{table}[h]
    \footnotesize
    \centering
    \begin{tabular}{c|ccc|cccc}
      \toprule
        & S1  & S2  & S3 & \textbf{S4} & \textbf{S5} & \textbf{S6} & \textbf{S7}  \\
      \midrule
        B1 & 5.1 & 5.7 & 5.0 &  5.1 & 4.9 & 4.5 & 3.9 \\
        B2 & 6.5 & 7.0 & 6.0 &  7.6 & 7.6 & 7.5 & 6.8 \\
        B3 & 5.9 & 6.7 & 5.8 &  8.0 & 8.1 & 8.2 & 7.9 \\
        \midrule
        \textbf{B4} & 6.4 & 7.1 & 5.9 &  7.5 & 7.6 & 7.3 & 7.3 \\
        \textbf{B5} & 6.3 & 7.1 & 6.2 &  6.8 & 6.8 & 6.4 & 6.2 \\
        \textbf{B6} & 6.2 & 7.1 & 6.1 &  7.8 & 7.8 & 7.4 & 7.0 \\
        \textbf{B7} & 6.2 & 7.0 & 6.2 &  7.8 & 7.8 & 8.0 & 7.8 \\
        \midrule
        B8 & 4.3 & 5.3 & \textbf{5.4} &  3.1 & 2.8 & 3.0 & 3.0 \\
        B9 & 4.9 & 5.7 & 5.5 &  2.3 & 2.1 & 2.1 & 1.8 \\
      \bottomrule
    \end{tabular}
    \vspace{2mm}
    \caption{\footnotesize Self-play agent performance table $\mathcal{P}$. Each row is for one bait agent named Bi in $\mathcal L_\B$ (i=1 to 9), and each column is for one shooter agent named Sj in $\mathcal L_\S$ (j=1 to 7). Each entry is computed by per-minute team performance (number of fortress kills) of the corresponding pair. We segment the tables to group same type of agents, and mark the ``optimal" bait and shooter agents in bold.}
    \label{tab:perf}
\end{table}

\subsection{Similarity Metric}
\label{sec:sim}
Now we introduce the \textbf{cross-entropy metric} (CEM) as the similarity metric used in this architecture. 
Cross-entropy, well-known in information theory, can measure the (negative) distance between two policies $\pi_1,\pi_2$:
\begin{equation}
\label{eq:cem}
\mathrm{CEM}(\pi_1, \pi_2) \defeq \E{s,a \sim \pi_1}{\log \pi_2 (a | s)}
\end{equation}                  
where $\pi_1(\cdot |s), \pi_2(\cdot |s)$ are action distributions given state $s$. 
This is actually the training objective of behavior cloning~\cite{bain1995framework} to expert policy $\pi_1$, i.e., $\max_{\pi_2} \mathrm{CEM}(\pi_1, \pi_2)$, which is to maximize the log-likelihood of expert actions in agent policy $\pi_2$ given a collection of expert state-actions. Thus the larger the $\mathrm{CEM}(\pi_1, \pi_2)$, the more similar $\pi_1$ is to $\pi_2$. 

If we know the policy $\pi_2$, and are able to obtain state-action samples from $\pi_1$, then we can estimate cross-entropy $\mathrm{CEM}(\pi_1, \pi_2)$ by Monte Carlo sampling. That is to say, under the assumption above, policy $\pi_1$ can be unknown to us.
In human-agent teaming, human policy $\pi_H$ cannot be observed but the state-action pairs generated by the human policy can be easily obtained, and agent policy $\pi_A$ is designed by us, as programmers, thus known to us. 

Therefore, we can leverage CEM as the similarity metric: given a sliding window of frames that record the observed behavior of the human policy $\pi_H$, we can estimate the cross-entropy between a human policy $\pi_H$ and any known agent policy $\pi_A$ by the following formula:
\begin{equation}
\label{eq:cem_mc}
\frac{1}{T}\sum_{t=1}^T \log \pi_A (a_t | s_t), \quad \mathrm{where}\, (s_t,a_t)_{t=1}^T \sim \pi_H
\end{equation}
where $(s_t,a_t)_{t=1}^T$ are the sequential state-action pairs from human policy, $T$ is the window size, which is a hyperparameter to be tuned.

\subsection{Adaptive Agent Architecture}
\label{sec:framework}
The prerequisite for the architecture is the exemplar policies library $\mathcal L$ introduced in the Sec.~\ref{sec:lib} and the self-play table $\mathcal P$ of the library to translate human-agent performance in the adaptation process.

Figure~\ref{fig:flowchart} shows the overall flowchart of our adaptive agent framework. 
When the game starts and a new human player $A$ starts to play as one pre-specified role $R_1\in \{\B, \S\}$ in TSF, the adaptive agent framework will first randomly assign a policy $B$ from the library $\mathcal L_{R_2}$ in teammate role $R_2$ such that $\{R_1, R_2\} = \{\B, \S\}$, and keep track of the joint trajectories (state-action sequences) and record them into memory. 

The adaptation process is as follows. As we maintain the latest human trajectories of a pre-specified window size, and we first use the data to compute the similarity by cross-entropy metric between the human trajectory and any of exemplar policies in the library $\mathcal L_{R_1}$ with same role. 
Then we figure out the most similar policy $C\in \mathcal L_{R_1}$ to the human trajectory, and look up the performance table $\mathcal P$ to find the optimal complementary policy $D\in \mathcal L_{R_2}$ to the predicted human policy type $C$. Finally, we assign the agent $D$ as the complementary policy at next timestamp with the human player.

The adaptation process on the exemplar policies selection is based on the following assumption: if the human policy $A$ with role $R_1$ is similar to one exemplar policy $C \in \mathcal L_{R_1}$ within some threshold, then the human policy $A$ will have similar team performance with teammates as $C$, i.e., if $C$ performs better with $D\in \mathcal L_{R_2}$ than $E\in\mathcal L_{R_2}$, so does $A$. This enables us to adapt the agent policy in real-time by the recent data without modeling the human policy directly.

\section{Human-Agent Teaming Experiments}
In this section, we first introduce our experiment design for human-agent teaming, 
then evaluate the human-agent performance when paired with static policy agents (introduced in Sec.~\ref{sec:lib}) and proposed adaptive agents (introduced in Sec.~\ref{sec:framework}).

By analyzing the collected human-agent data, we aim to answer the following motivated questions:
\begin{enumerate}
    \item How are human players' policies compared to agent policies in our library?
    \item Is our adaptive agent architecture capable of identifying human policies and predicting team performance for human-agent teams?
    \item Do our adaptive agents perform better than static policy agents in human-agent teams?
\end{enumerate}

\subsection{Experimental Design}
We recruited participants from Amazon Mechanical Turk for our human-agent experiments. They were paid USD 2 for participating in the 15-min online study. Participants were randomly assigned a role of either shooter or bait and then teamed with artificial agents in the corresponding role to play Team Space Fortress. Each participant would need to complete five sessions of data collection with three 1-min game trials in each session. Participants teamed with different agent variants between sessions in a random sequence. The five variants were selected from our static agent library $\mathcal{L}$. When selecting these designated agents, we balanced the performance in self-play table and the diversity by considering different training methods and reward functions. Specifically, we select $\{$B3, B6, B7, B8, B9$\}$ as tested static baits and $\{$S1, S2, S3, S4, S7$\}$ as tested static shooters. In the dataset of human and static agent teams, we got 25 valid data points from human shooters and 29 valid data from human baits.

\subsection{Results}

\subsubsection{Policy space representation}
\label{sec:q1}

To quantify the relationship between real human policies in the experiments and agent policies in the library, we leveraged a similarity embedding by comparing the distance between the collected human trajectories and agent policies using CEM measurement (see Sec.~\ref{sec:sim}). This provides us with a high-dimensional policy space based on agent policies in our library. 
Specifically, CEM was employed to generate the average log-probability of state-action pairs in a human trajectory coming from a certain agent policy. We could then construct a similarity vector for each trajectory with the dimensions equal to the number of policies in the library. The value in each dimension represents the similarity distance from human trajectories to a certain agent policy. 
Then, we applied a principal component analysis (PCA) based on the log-probability dataset to project the high-dimensional policy space into a 2D plane for a better visualization. The two primary components left explain more than 99\% of the variance. 

\begin{figure}[htbp]
    \centering
    \includegraphics[width=0.48\textwidth]{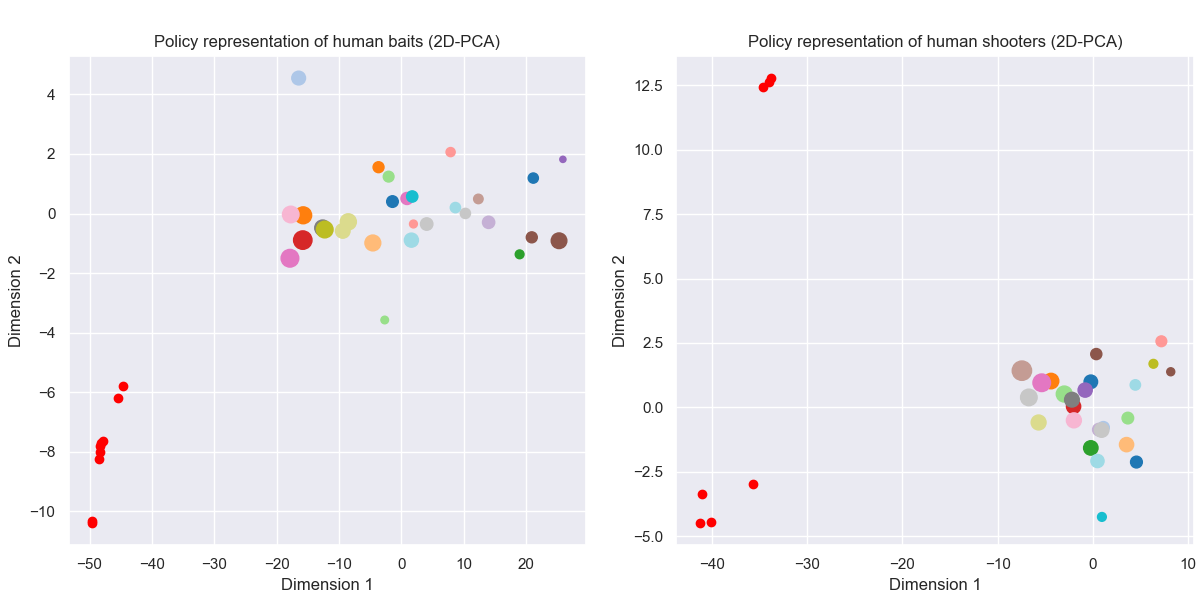}
    \caption{\footnotesize Policy representations of each human baits (left) and shooters (right) in the static agent dataset (after PCA dimension reduction). 
    Each colored node in the figures represents the average policy of a human player, while the size of which indicates his average team performance. Red nodes are reference points of baseline agent polices.
    }
    \label{fig:policy_space}
\end{figure}

Fig.~\ref{fig:policy_space} illustrates the human policies in static agent dataset. 
We could get following qualitative insights from the illustration: 1) the learnt similarity embedding separates different human policies well, 2) reinforcement learning policies are homogeneous (red nodes in the bottom-left corner) while the rule-based policies are a bit off (red nodes in the upper-left corner). 3) the distribution of human policies correlates with their team performance in that players to the left tend to have better team performance (colored nodes to the left are larger in size). 
Those findings align with our expectations and validate the proposed adaptive agent architecture. In the following analysis, we will quantify them based on the CEM measurement and the similarity embedding.

\subsubsection{Human policy identification}
\label{sec:q2}

In the proposed adaptive agent architecture, our model infers human policy by classifying it as the most similar policy in the library based on CEM measurement, then assigns the agent with the corresponding complementary policy in the self-play table. One way of verifying this method is to see if human-agent teams performed better when the predicted human policy was closer to the complementary match in the self-play table $\mathcal{P}$. Assuming each human maintains a consistent policy over the course of interaction when paired with a specific teammate with static policy, we could then calculate, for each human-agent pair, the similarity between human policy and the optimal agent policy for the agent that the human was playing with. 

This ``similarity to optimal" quantifies the degree to which a human player is similar to the optimal policy given an agent teammate in our architecture. Correlation analysis shows that ``similarity to optimal" is positively correlated with team performance in both bait ($r = 0.636, p =.0002$) and shooter ($r = 0.834, p <.0001$) groups. 
This result indicates that the complementary policy pairs we found in agent-agent self-play can be extended to human-agent teams, and our proposed architecture is able to accurately identify human policy types and predict team performance. 

Furthermore, our model could also infer human policies in real-time. This is to say, even within the same team, humans might also take different sub-policies as their mental model of the team state evolves over time~\cite{salas2008teams}. We could take the log-probabilities generated by CEM as time series data to capture the online adaptation process of humans over the course of interaction at each timestamp. 

\begin{figure}[htbp]
    \centering
    \includegraphics[width=0.48\textwidth]{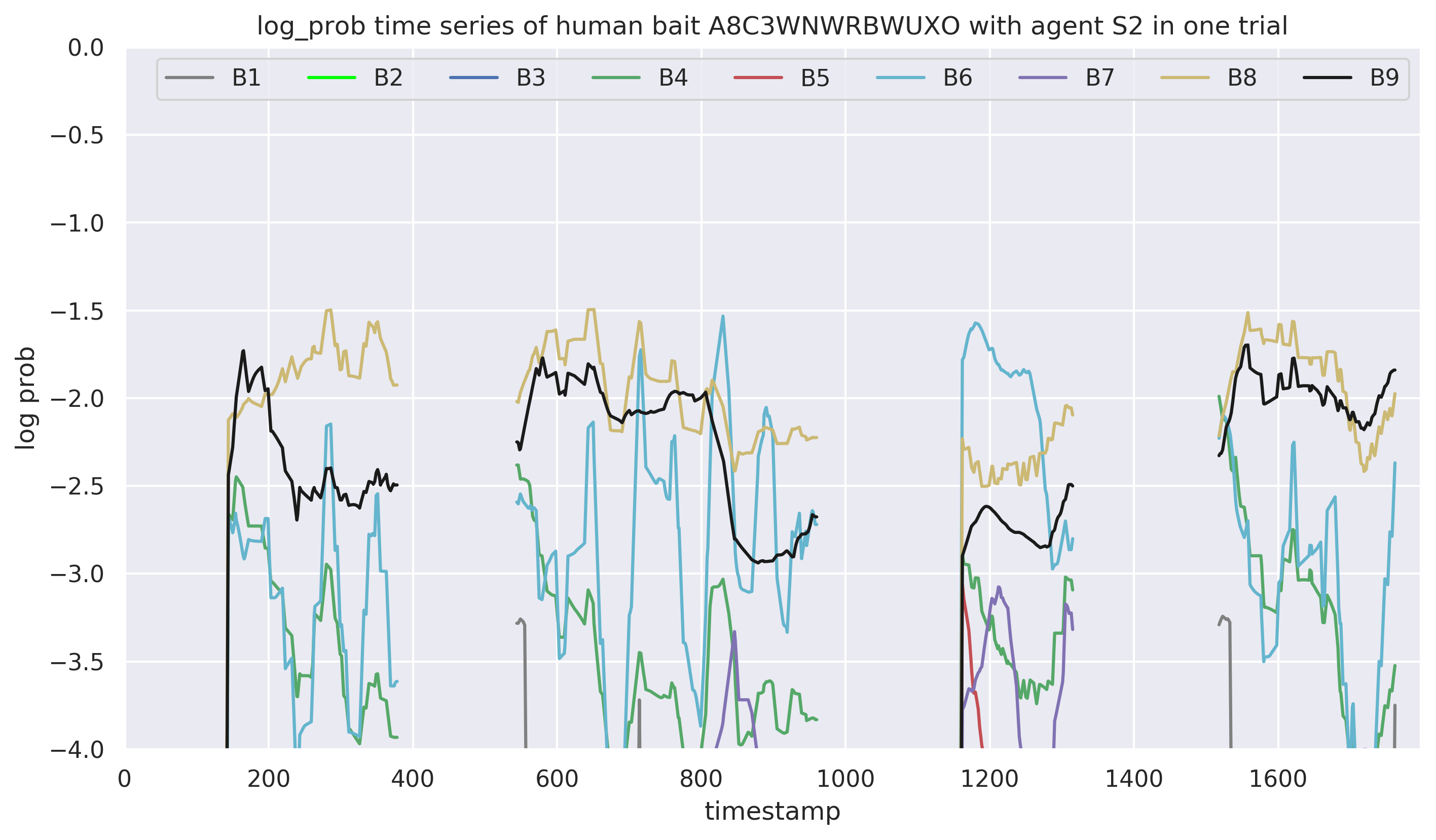}
    \caption{\footnotesize The log-probability curves of one human policy generated by CEM. Data is from one specific 1-min trial of a human bait paired with agent S2. We segment the trial into several episodes, each of which starts with the bait entering the activation region and ends with the team killing the fortress. The curves with same color represent the same agent policy for inference.}
    \label{fig:log_prob}
\end{figure}

An example visualization is shown in Fig.~\ref{fig:log_prob} where curves represent the log-probabilities of each agent policy over the course of interaction. We can tell from the graph that in the segments of a specific trial, the human trajectories were inferred to reflect different policies, although the average log-probability would still be in favor of B8. Those findings motivate us to test an online adaptive agent using a sliding time window to capture the human policy shifts in real-time.

\subsection{Pilot experiment with adaptive agents}

\label{sec:q3}

In previous experiment and analysis, we validated our proposed architecture on the static agent dataset. Finally, we conducted a pilot experiment to measure the performance of adaptive agents in HATs by pairing them with human players. 

In the adaptive agent experiment, adaptive agent uses the CEM similarity metric (Sec. ~\ref{sec:sim}) to identify the policy most similar to the human behavior over a fixed number of recent preceding game frames. The frames were tracked using a sliding window, the size of which was adjusted during the hyperparameter tuning phase of experimentation.
To perform the adaptation procedure, in each frame, after identifying the most similar agent to the human teammate, the agent referenced the self-play table to select the policy that would best complement the teammate's estimated policy.

In this round of experiment, the five variants including three values of the window size hyperparameter ($T$ in Eq.~\ref{eq:cem_mc}) for the adaptive agent (150, 400, 800 frames) and two best-performed static agent policy (representing the extreme condition of 0 window size where the adaptive agent becomes static). Besides that, all experimental settings are the same as in static agent experiment. We got in total, 22 valid data points from human shooters and 25 valid data from human baits.

\begin{figure}[htbp]
    \centering
    \includegraphics[width=0.48\textwidth]{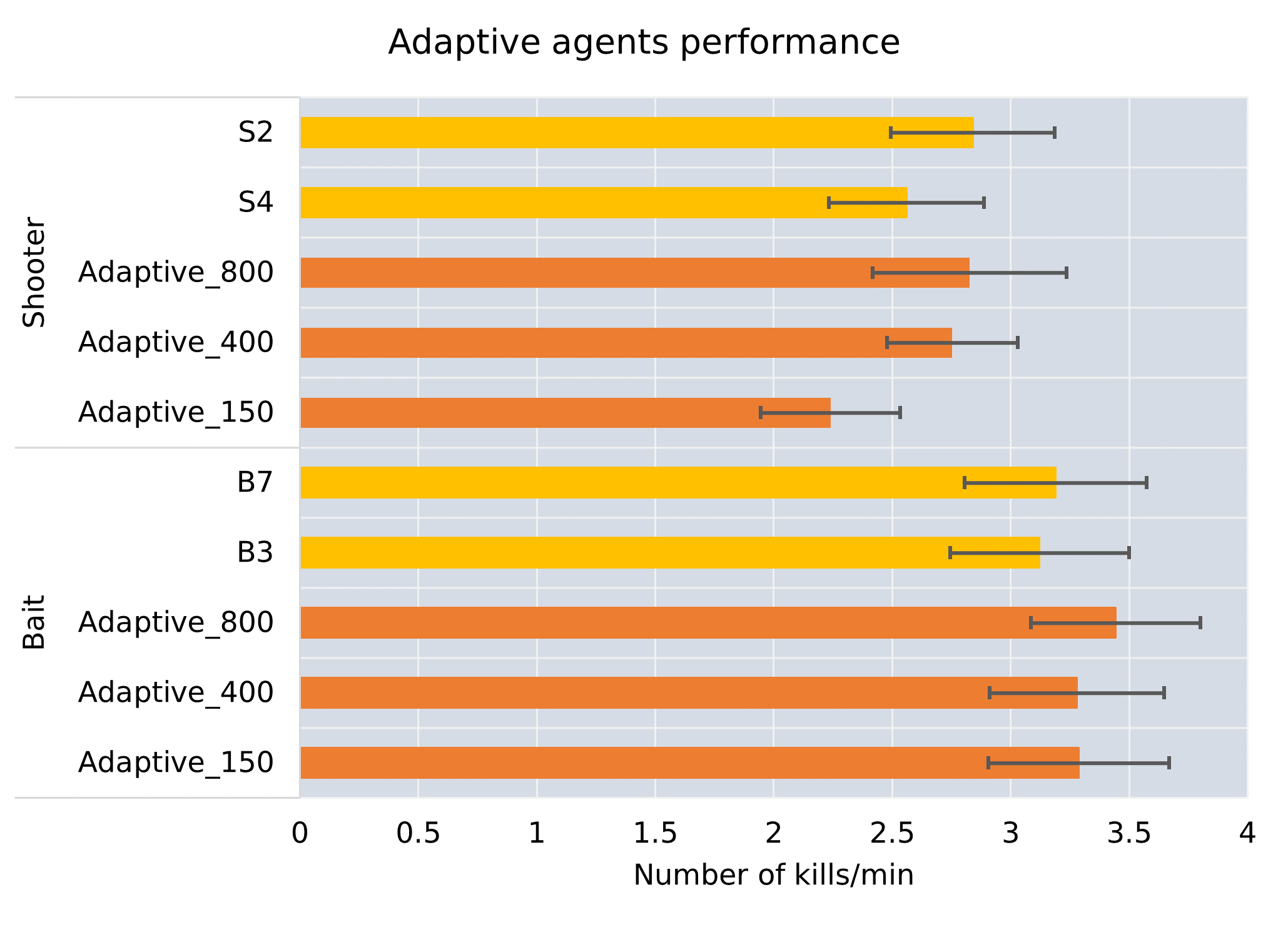}
    \caption{\footnotesize Human-agent team performance when humans paired with adaptive or static agent policies. Error bars represent one standard error away from the mean.}
    \label{fig:agent_per}
\end{figure}

Fig.~\ref{fig:agent_per} shows the average team performance of HATs when human players were paired with either static or adaptive agents. We could see from the figure that adaptive agents (marked in orange) have slightly better performance than static agents (marked in yellow), although not statistically significant. 
In addition, adaptive agents with longer time window (e.g. 800 frames) tend to have better performance in HATs since they accumulate more evidence for human policy inference. However, a larger sample size and and better hyper-parameter tuning might be necessary for future research to confirm the advantage of adaptive agents in HATs.

\section{Conclusion and Future Work}

In this paper, we proposed a novel adaptive agent framework in human agent teaming (HAT) based on the cross-entropy similarity measure and a pre-trained static policy library. 
The framework was inspired by human teamwork research, which illustrates important characteristics of teamwork such as the existence of complementary policies, influence of adaptive actions on team performance, and the dynamic human policies in cooperation~\cite{li2020individual,li2020team}. 
Those findings motivate us to introduce an online adaptive agents into HATs in order to maximize the team performance even when given unknown human teammates. 
The proposed framework adopts a human-model-free method to reduce the computational cost in real-time deployment and make the pipeline more generalizable to diverse task settings and human constraints. 

The specific task scenario studied in this paper, i.e. Team Space Fortress, is a nontrivial cooperative game which requires sequential decision-making and real-time cooperation with heterogeneous teammates. We evaluated the validity of proposed adaptive agent framework by running human-agent experiments. 
Results show that our adaptive agent architecture is able to identify human policies and predict team performance accurately. We constructed a high-dimensional policy space based on exemplar policies in a pre-trained library and leveraged it as a standard and reliable way to categorize and pair human policies. The distance between human policy and the optimal complementary for his/her teammate is shown to be positively correlated with team performance, which confirms the validity of our proposed framework. In additional, we found that human players showed diverse policies in HAT (1) when paired with different teammates (2) over the course of interaction within the same team. These findings point out that we cannot simply impose strong assumptions on humans, e.g. optimality, consistency, and unimodality, prevalent in human-model-based settings. Thus, we employed an online inference mechanism to identify the human policy shifting during the course of interaction and adapt the agent policy in real time.

As for future directions, we would like to enrich the static agent library by introducing novel policies such as imitation learning agents that learn from human demonstrations. A larger coverage in the policy space of exemplar policies library could lead to a more accurate estimation of human policy and a better selection of complementary policy.

\section*{Acknowledgments}
This research was supported by a reward W911NF-19-2-0146 and AFOSR/AFRL award FA9550-18-1-0251.

\bibliography{refs}

\end{document}